\documentclass{article}
\usepackage{ijcai21}

\usepackage{times}
\usepackage{soul}
\usepackage{url}
\usepackage[hidelinks]{hyperref}
\usepackage[utf8]{inputenc}
\usepackage[small]{caption}
\usepackage{booktabs}
\usepackage{graphicx}
\usepackage{amsmath}
\usepackage{subcaption}
\usepackage{amsthm}
\usepackage{booktabs}
\usepackage{algorithm}
\usepackage{algorithmic}
\usepackage{multirow}
\usepackage{mathtools}
\usepackage{array}
\usepackage{enumitem}
\usepackage{setspace}
\usepackage{amssymb}
\usepackage{amsfonts}
\newcommand{\Eb}[2]{\E_{#1}\!\left[#2\right]}
\newcommand{\E}{\mathbb{E}}
\newcommand{\etalcite}[1]{\textit{et al.}~\shortcite{#1}}
\usepackage{array, multirow}
\newcolumntype{C}[1]{>{\centering\let\newline\\\arraybackslash\hspace{0pt}}m{#1}}

\title{SRDiff: Single Image Super-Resolution with Diffusion Probabilistic Models}

\author{
Haoying Li$^1$\and
Yifan Yang$^{1}$\and
Meng Chang$^{1}$\and
Huajun Feng$^1$\footnote{Corresponding Author}\and 
Zhihai Xu$^{1}$\and
Qi Li$^{1}$\And
Yueting Chen$^1$\\
\affiliations
$^1$Zhejiang University\\
\emails
\{lhaoying,yangyifan,changm,fenghj,xuzhi,liqi,chenyt\}@zju.edu.cn
 }

\begin{document}

\maketitle

\begin{abstract}
Single image super-resolution (SISR) aims to reconstruct high-resolution (HR) images from the given low-resolution (LR) ones, which is an ill-posed problem because one LR image corresponds to multiple HR images. Recently, learning-based SISR methods have greatly outperformed traditional ones, while suffering from over-smoothing, mode collapse or large model footprint issues for PSNR-oriented, GAN-driven and flow-based methods respectively. To solve these problems, we propose a novel single image super-resolution diffusion probabilistic model (SRDiff), which is the first diffusion-based model for SISR. SRDiff is optimized with a variant of the variational bound on the data likelihood and can provide diverse and realistic SR predictions by gradually transforming the Gaussian noise into a super-resolution (SR) image conditioned on an LR input through a Markov chain. In addition, we introduce residual prediction to the whole framework to speed up convergence. Our extensive experiments on facial and general benchmarks (CelebA and DIV2K datasets) show that 1) SRDiff can generate diverse SR results in rich details with state-of-the-art performance, given only one LR input; 2) SRDiff is easy to train with a small footprint; and 3) SRDiff can perform flexible image manipulation including latent space interpolation and content fusion. 
\end{abstract}

\section{Introduction}
\begin{figure}[h]
	\begin{center}
		\includegraphics[width=0.5\textwidth,trim={0cm 0cm 0.5cm 0cm}, clip=true]%
		{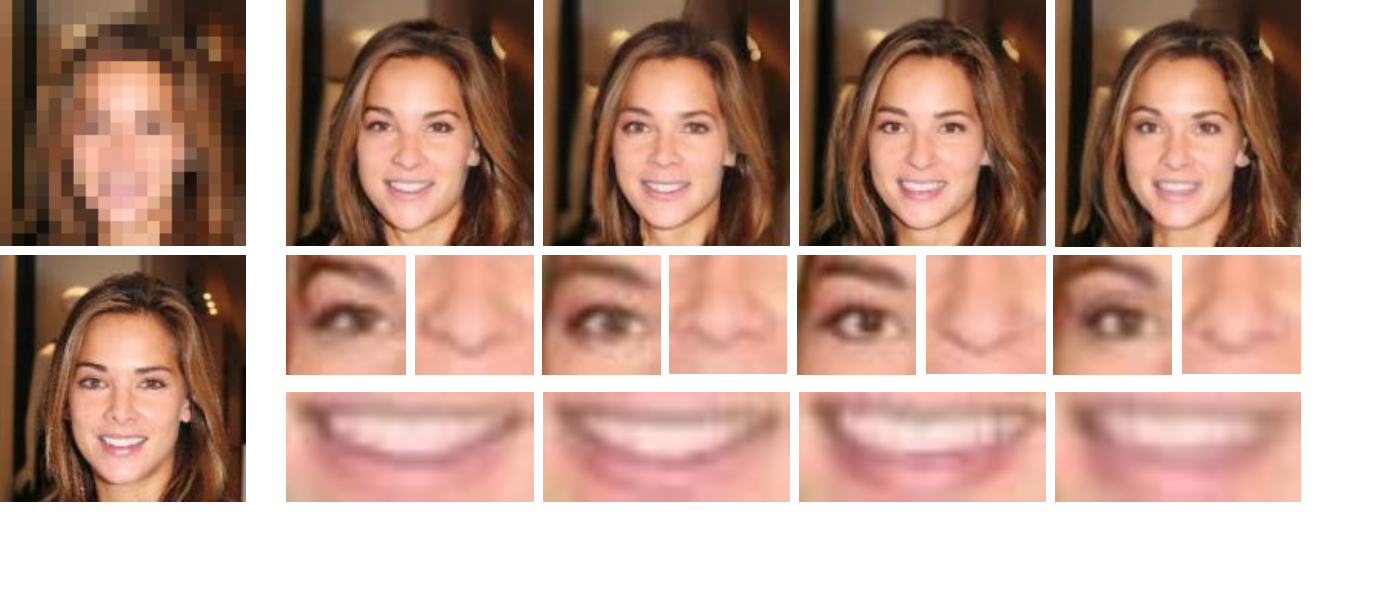}
	\end{center}
	\caption{Random SR (8x) predictions generated by our method. LR/HR image is shown on top top/bottom left. The right columns are diverse SR predictions and their facial regions, which are different from each other in expressions and attributes. For example, the first nose looks flat, while the third looks straight.}
	\label{fig:faces_diverse}
\end{figure}

Over the years, single image super-resolution (SISR) has drawn active attention due to its wide applications in computer vision such as object recognition \cite{fookes2012evaluation,sajjadi2017enhancenet}, remote sensing \cite{li2009super}, surveillance monitoring \cite{fang2019novo,park2020self} and so on. SISR aims to recover high-resolution (HR) images from low-resolution (LR) ones, which is an ill-posed problem, for multiple HR images can be degenerated to one LR image as shown in Figure \ref{fig:faces_diverse}. 

To establish the mapping between HR and LR images, lots of deep learning-based methods emerge and could be categorized into three main types: PSNR-oriented, GAN-driven and flow-based methods. PSNR-oriented methods \cite{dong2015image,lim2017enhanced,zhang2018image,qiu2019embedded,guo2020closed} are trained with simple distribution assumption-based losses (e.g., Laplacian for $L_1$ and Gaussian for $L_2$) and achieve excellent PSNR. However, these losses tend to drive the SR result to an average of several possible SR predictions, causing \textbf{over-smoothing} images with high-frequency information loss. One ground-breaking solution to tackle the over-smoothing problem is GAN-driven methods \cite{cheon2018generative,kim2019progressive,ledig2017photo,wang2018esrgan}, which combine content losses (e.g., $L_1$ and $L_2$) and adversarial losses to obtain sharper SR images with better perceptual quality. However, GAN-driven methods are easy to fall into \textbf{mode collapse}, which leads to a single generated SR sample without diversity. Additionally, GAN-based training process is not easy to converge and requires an extra discriminator which is not used in inference. Flow-based methods~\cite{lugmayr2020srflow} directly account for the ill-posed problem with an invertible encoder, which maps HR images to the flow-space latents conditioned on LR inputs. Trained with a negative log-likelihood loss, flow-based methods avoid training instability but suffer from \textbf{extremely large footprint and high training cost} due to the strong architectural constraints to keep the bijection between latents and data. 

Lately, the successful adoptions of diffusion probabilistic models (diffusion models for short) in image synthesis \cite{ho2020denoising} and speech synthesis \cite{kong2020diffwave,chen2020wavegrad} witness the power of diffusion models in generative tasks. The diffusion models use a Markov chain to convert data $x_0$ to latent variable $x_T$ in simple distribution (e.g., Gaussian) by gradually adding noise $\epsilon$ in the diffusion process, and predict the noise $\epsilon$ in each diffusion step to recover the data $x_0$ through a learned reverse process. Diffusion models are trained by optimizing a variant of the variational lower bound, which is efficient and avoids the mode collapse encountered by GAN.

In this paper, we propose a novel single image super-resolution diffusion probabilistic model (SRDiff) to tackle the over-smoothing, mode collapse and huge footprint problems in previous SISR models. Specifically, 1) to extract the image information in LR image, SRDiff exploits a pretrained low-resolution encoder to convert LR image into hidden condition. 2) To generate the HR image conditioned on LR image, SRDiff employs a conditional noise predictor to recover $x_0$ iteratively. 3) To speed up convergence and stabilize training, SRDiff introduces residual prediction by taking the difference between the HR and LR image as the input $x_0$ in the first diffusion step, making SRDiff focus on restoring high-frequency details. To the best of our knowledge, SRDiff is the first diffusion-based SR model and has several advantages:  
\begin{itemize}
\item \textbf{Diverse and high-quality outputs}: SRDiff converts Gaussian white noise into an SR prediction through a Markov chain, which does not suffer from mode collapse and can generate diverse and high-quality SR results.  
\item \textbf{Stable and efficient training with small footprint}: Although the data distribution of HR image is hard to estimate, SRDiff admits a variant of the variational bound maximization and applies residual prediction. Compared with GAN-driven methods, SRDiff is stably trained with a single loss and does not need any extra module (e.g., discriminator, which is only used in training). Compared with flow-based methods, SRDiff has no architectural constraints and thus enjoys benefits from small footprint and fast training.
\item \textbf{Flexible image manipulation}: SRDiff can perform flexible image manipulation including latent space interpolation and content fusion using both diffusion process and reverse process, which shows broad application prospects. 
\end{itemize} 
Our extensive experiments on CelebA~\cite{liu2015deep} and DIV2K~\cite{timofte2018ntire} datasets show that 1) SRDiff can reconstruct multiple SR results given one LR input and outperform state-of-the-art SISR methods; 2) SRDiff only has 1/4 parameters and is stable and fast to train (about 30 hours on 1 GPU until convergence) compared with SRFlow; 3) we can manipulate the generated SR images in latent space to obtain more diverse outputs. 

\section{Related Works} 
\subsection{Single Image Super-Resolution}
Recently, deep learning methods have become widely adopted to single image super-resolution (SISR) and we categorize them into three types: PSNR-oriented and GAN-driven and flow-based methods. The training goal of \textbf{PSNR-oriented methods} is to minimize the mean squared error (MSE) between the ground truth and the SR image recovered from the LR image. SRCNN \cite{dong2015image} sets a precedent of end-to-end mapping between the LR and HR images. Kim \etalcite{kim2016deeply,kim2016accurate} then apply residual neural network techniques to SR tasks and deepen the network. Some works\cite{lim2017enhanced,zhang2018image,qiu2019embedded,guo2020closed} enhance the SR performance by carefully adjusting network structures and losses. \textbf{GAN-driven methods} \cite{cheon2018generative,kim2019progressive} solve the over-smoothing problem towards perceptual restrictions \cite{rad2019srobb,zhang2019ranksrgan}. The pioneer work is SRGAN \cite{ledig2017photo}, using SRResNet \cite{ledig2017photo} and perceptual loss as well as the adversarial loss to improve the naturalness of the recovered image. ESRGAN \cite{wang2018esrgan} further enhances it with network adjustments of structure and loss function. The first \textbf{flow-based methods} is SRFlow~\cite{lugmayr2020srflow}, which builds an invertible neural network to transform a Gaussian distribution into an HR image space instead of modeling one single output and inherently resolves the pathology of the original "one-to-many" SR problem. 

\subsection{Diffusion models} 
Diffusion probabilistic models \cite{sohl2015deep,ho2020denoising} are a kind of generative models using a Markov chain to transform latent variables in simple distributions (e.g., Gaussian) to data in complex distributions. Researchers find it useful to tackle "one-to-many" problems and synthesize high-quality results in speech synthesis tasks \cite{kong2020diffwave} and image synthesis fields \cite{ho2020denoising}. However, to the best of our knowledge, diffusion models have not yet been used in image reconstruction fields like super-resolution. In this paper, we propose our impressive work, SRDiff, building on diffusion models to generate diverse SR images with a single LR input, and solving over-smoothing, mode collapse and large footprint issues together.

\section{Diffusion Model} 
In this section, to provide a basic understanding of diffusion probabilistic models (diffusion model for short) \cite{ho2020denoising}, we first briefly review its formulation.

\begin{figure}[h]  
\centering 
\includegraphics[width=0.48\textwidth,trim={0cm 0.8cm 1.3cm 0cm}, clip=true]{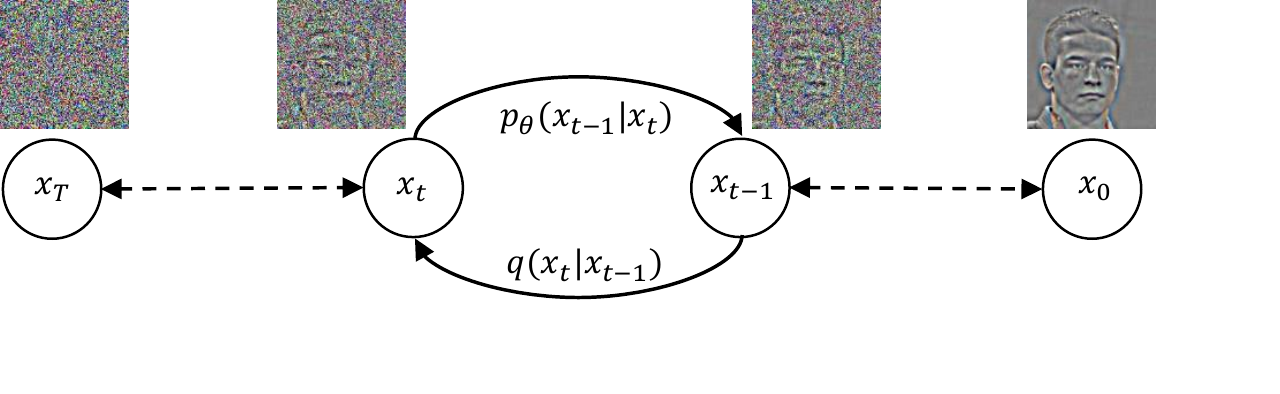} 
\caption{Overview of two processes in SRDiff. The diffusion process is from right to left and the reverse process is from left to right. $\theta$ in $p_\theta$ denotes the learnable components including conditional noise predictor and low-resolution encoder in SRDiff.} 
\label{fig:srdiff_processes} 
\end{figure}

\begin{figure*}[t]  
\centering 
\includegraphics[width=1\textwidth,trim={0cm 1.6cm 1.3cm 0.2cm}, clip=true]{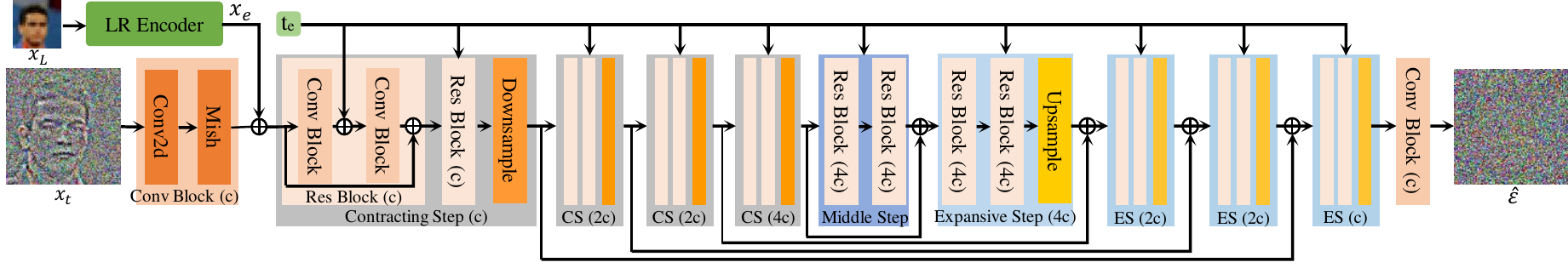} 
\caption{The architecture of conditional noise predictor in SRDiff. The content in parentheses (c, 2c and 4c) after the block name means the channel size of each block. ``Conv Block", ``Res Block", ``Downsample" and ``Upsample" denote 2D-Convolution block, residual block, downsampling layer and upsampling layer respectively; ``CS" and``ES" denote ``Contracting Step" and ``Expansive Step" respectively.} 
\label{fig:conditional_noise_predictor} 
\end{figure*} 

\begin{figure*}[th]
\begin{minipage}[t]{0.495\textwidth}
\begin{algorithm}[H] 
    \small
    \centering 
    \caption{Training}\label{alg: training} 
    \begin{algorithmic}[1] 
    \STATE\textbf{Input}: LR image and its corresponding HR image pairs $P=\{(x_L^k,~x_H^k)\}_{k=1}^K$, total diffusion step $T$
    \STATE\textbf{Initialize}: randomly initialized conditional noise predictor $\epsilon_\theta$ and pretrained LR encoder $\mathcal{D}$ 
    \REPEAT 
        \STATE Sample $(x_L,~x_H)\sim P$
        \STATE Upsample $x_L$ as $up(x_L)$, compute $x_r=x_H-up(x_L)$
        \STATE Encode LR image $x_L$ as $x_e = \mathcal{D}(x_L)$ 
        \STATE Sample $\epsilon\sim\mathcal{N}(\textbf{0},\textbf{I})$, and $t\sim\mathrm{Uniform}(\{1,\cdots,T\})$
        \STATE Take gradient step on \\ 
        \quad $\nabla_{\theta}\|\epsilon-\epsilon_{\theta}(x_t,x_e,t)\|$, $x_t=\sqrt{\bar{\alpha}_t}x_r+\sqrt{1-\bar{\alpha}_t}\epsilon$ \\ 
    \UNTIL{converged} 
    \end{algorithmic} 
\end{algorithm} 
\end{minipage}
\hfill
\begin{minipage}[t]{0.495\textwidth}
\begin{algorithm}[H] 
    \setstretch{1.045}
    \centering 
    \small
    \caption{Inference}\label{alg: sampling} 
    \begin{algorithmic}[1] 
        \STATE\textbf{Input}: LR image $x_L$, total diffusion step $T$
        \STATE\textbf{Load}: conditional noise predictor $\epsilon_\theta$ and LR encoder $\mathcal{D}$
        \STATE Sample $x_T\sim\mathcal{N}(\textbf{0},\textbf{I})$
        \STATE Upsample $x_L$ to $up(x_L)$ 
        \STATE Encode LR image $x_L$ as $x_e = \mathcal{D}(x_L)$ 
        \FOR{$t=T,T-1,\cdots,1$} 
            \STATE Sample $z \sim \mathcal{N}(\textbf{0},\textbf{I})$ if $t > 1$, else $z = 0$ 
            \STATE Compute $x_{t-1}$ using Eq. 
            \eqref{eq:parameterization}: \\ 
            $x_{t-1} = \frac{1}{\sqrt{\alpha_t}}\left(x_t-\frac{1-\alpha_t}{\sqrt{1-\bar\alpha_t}} \epsilon_\theta(x_t,x_e,t)\right) + \sigma_\theta(x_t, t) z$ 
        \ENDFOR 
        \RETURN{$x_0 + up(x_L)$} as SR prediction 
    \end{algorithmic} 
\end{algorithm} 
\end{minipage}
\end{figure*}

A diffusion model is a kind of generative model which adopts parameterized Markov chain trained using variational inference to gradually generate data $x_0$ in complex distribution from a latent variable $x_T$ in simple distribution, where $T$ is the total diffusion step. We set $x_t \in \mathbb{R}^d$ to be the results of each diffusion timestep $t\in \{1, 2, ... ,T\}$ and  $x_t$ shares the same dimension $d$ as that of $x_0$. As shown in Figure \ref{fig:srdiff_processes}, a diffusion model is composed of two processes: the \textbf{diffusion process} and the \textbf{reverse process}. 

The posterior $q(x_1,\cdots,x_T|x_0)$, named as the \textbf{diffusion process}, converts the data distribution $q(x_0)$ to the latent variable distribution $q(x_T)$, and is fixed to a Markov chain which gradually adds Gaussian noise $\epsilon$ to the data according to a variance schedule $\beta_1,\cdots,\beta_T$: 
\begin{align} 
q(x_1,\cdots,x_T|x_0) & := \prod_{t=1}^T q(x_t|x_{t-1}), \nonumber \\ 
q(x_t|x_{t-1}) & := \mathcal{N} (x_t;\sqrt{1-\beta_t}x_{t-1},\beta_t \textbf{I}), \nonumber  
\end{align} 
where $\beta_t$ is a small positive constant and could be regarded as constant hyperparameters. Setting $\alpha_t := 1-\beta_t, ~ \bar{\alpha}_t := \prod_{s=1}^t\alpha_s$, the diffusion process allows sampling $x_t$ at an arbitrary timestep $t$ in closed form:  
\begin{align} 
  q(x_t|x_0) = \mathcal{N}(x_t; \sqrt{\bar\alpha_t}x_0, (1-\bar\alpha_t)\textbf{I}),
  \label{eq:q_marginal_arbitrary_t} 
\end{align} 
which can be further reparameterized as 
\begin{align} 
x_t(x_0, \epsilon) = \sqrt{\bar\alpha_t}x_0 + \sqrt{1-\bar\alpha_t}\epsilon ,~ \epsilon \sim \mathcal{N}(\textbf{0}, \textbf{I}). \label{eq:x_arbitrary_t_on_x0} 
\end{align} 

The \textbf{reverse process} transforms the latent variable distribution $p_\theta(x_T)$ to the data distribution $p_\theta(x_0)$ parameterized by $\theta$. It is defined by a Markov chain with learned Gaussian transitions beginning with $p(x_T) = \mathcal{N} (x_T; \textbf{0}, \textbf{I})$: 
\begin{align} 
p_{\theta}(x_0,\cdots,x_{T-1}|x_T) & := \prod_{t=1}^T p_{\theta}(x_{t-1}|x_t), \nonumber \\ 
p_\theta(x_{t-1}|x_t) & := \mathcal{N}(x_{t-1}; \mu_\theta(x_t, t), \sigma_\theta(x_t, t)^2 \textbf{I}),
\label{eq:reverse} 
\end{align} 

In training phase, we maximize the variational lower bound (ELBO) on negative log likelihood and introduce KL divergence and variance reduction \cite{ho2020denoising}: 
\begin{align} 
&\mathbb{E}[-\log p_\theta(x_0)] \leq L:=\mathbb{E}_q \bigg[ \underbrace{D_{KL}({q(x_T|x_0)}~||~{p(x_T)}}_{L_T}) \nonumber \\ 
&+ \sum_{t > 1} \underbrace{D_{KL}({q(x_{t-1}|x_t,x_0)}~||~{p_\theta(x_{t-1}|x_t))}}_{L_{t-1}} 
\underbrace{-\log p_\theta(x_0|x_1)}_{L_0} \bigg]. \label{eq:vb} 
\end{align} 
This transformation requires a direct comparison between $p_\theta(x_{t-1}|x_t)$ and its corresponding diffusion process posteriors. Setting $ \tilde\mu_t(x_t, x_0) := \frac{\sqrt{\bar\alpha_{t-1}}\beta_t }{1-\bar\alpha_t}x_0 + \frac{\sqrt{\alpha_t}(1- \bar\alpha_{t-1})}{1-\bar\alpha_t} x_t$, we have 
\begin{align} 
q(x_{t-1}|x_t,x_0) =  \mathcal{N}(x_{t-1}; \tilde\mu_t(x_t, x_0), \tilde\beta_t \textbf{I}). \label{eq:q_posterior_mean_var}
\end{align} 
Eq. \eqref{eq:q_marginal_arbitrary_t}, \eqref{eq:reverse} and \eqref{eq:q_posterior_mean_var} assure that all KL divergences in Eq. \eqref{eq:vb} are comparisons between Gaussians. With $\sigma_\theta^2=\tilde{\beta}_t=\frac{1-\bar{\alpha}_{t-1}}{1-\bar{\alpha}_t}\beta_t$ for $t>1,~ \tilde{\beta}_1=\beta_1$, and constant $C$, we have 
\begin{align} 
  L_{t-1} 
   = \Eb{q}{ \frac{1}{2\sigma_t^2} \|\tilde\mu_t(x_t,x_0) - \mu_\theta(x_t, t)\|^2 } + C. \nonumber
\end{align} 
For simplicity, the training procedure minimizes the variant of the ELBO with $x_0$ and $t$ as inputs: 
\begin{equation} 
\label{eq:train_obj} 
\min_{\theta} L_{t-1}(\theta) =  
\Eb{x_0,\epsilon,t}{ 
\|\epsilon-\epsilon_{\theta}(\sqrt{\bar{\alpha}_t}x_0+\sqrt{1-\bar{\alpha}_t}\epsilon,~t)\|^2},
\end{equation} 
where $\epsilon_{\theta}$ is a noise predictor. 

In inference, we first sample an $x_T \sim \mathcal{N} (x_T;\textbf{0}, \textbf{I})$, and then sample $x_{t-1} \sim p_\theta (x_ {t-1}|x_t)$ according to Eq. \eqref{eq:reverse}, where 
\begin{align} 
\mu_{\theta}(x_t, t) & := \frac{1}{\sqrt{\alpha_t}}\left(x_t-\frac{\beta_t}{\sqrt{1-\bar{\alpha}_t}}\epsilon_{\theta}(x_t, t)\right), \nonumber \\ 
\sigma_{\theta}(x_t, t) & := \tilde{\beta}_t^{\frac12},t\in\{T, T-1, ...,1\}. 
\label{eq:parameterization} 
\end{align} 

\begin{figure*}[t]
	\begin{center}
		\includegraphics[width=1.0\textwidth,trim={0cm 0.5cm 1.0cm 0cm}, clip=true]%
		{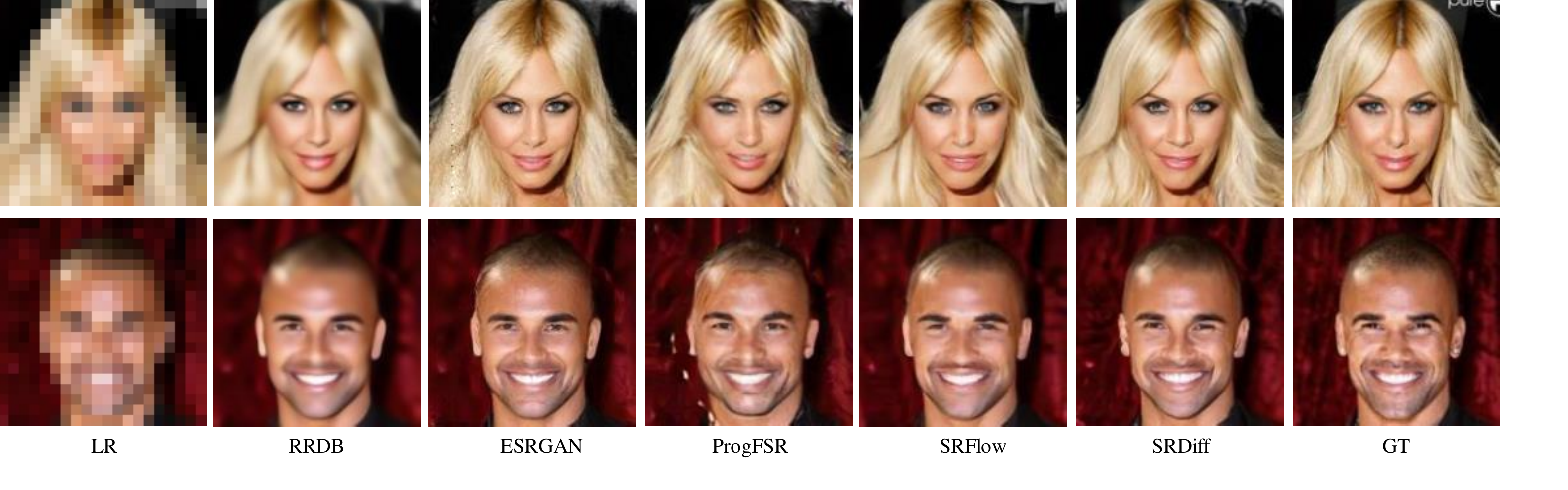}
	\end{center}
	\caption{Face SR $8\times$ visual results. SRDiff generates rich details than RRDB and SRFlow, avoids artifacts (e.g., grids on the woman's hair and stripes on man's head) encountered by ESRGAN and ProgFSR and maintain consistency with the ground truth.}
	\label{fig:faces8x}
\end{figure*}
\begin{figure*}[t]
	\begin{center}
		\includegraphics[width=1.0\textwidth,trim={0cm 1cm 2.6cm 0cm}, clip=true]%
		{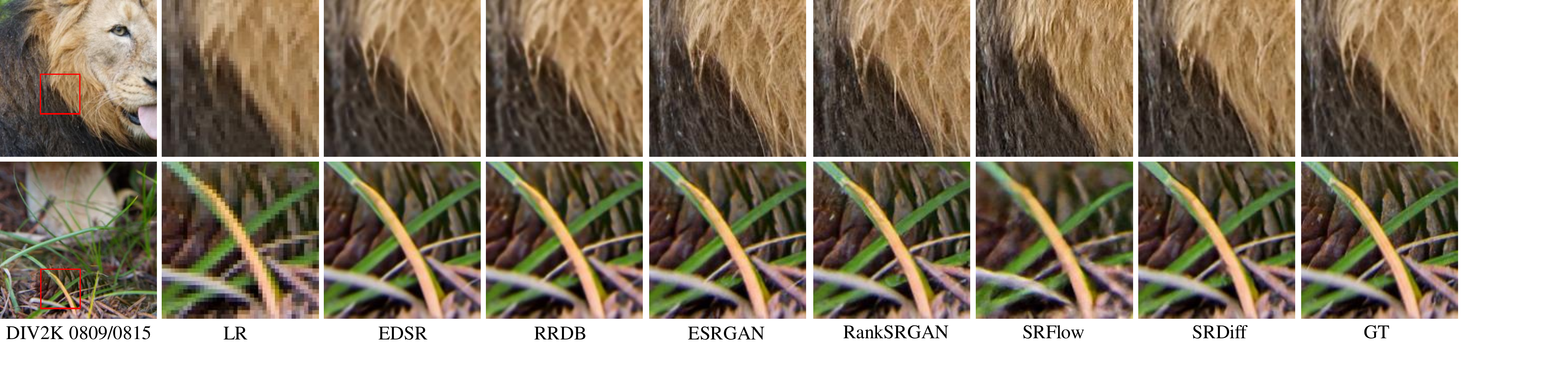}
	\end{center}
	\caption{General SR ($4\times$) visual results. SRDiff produces more natural textures, e.g., animal fur, and grass texture, and also less unpleasant artifacts, e.g., speckles on the yellow grass by ESRGAN and RankSRGAN. Only SRDiff maintains the brown streak on the yellow grass, which is consistent with the ground truth.}
	\label{fig:general4x}
\end{figure*}

\section{SRDiff} 
As depicted in Figure \ref{fig:srdiff_processes}, SRDiff is built on a T-step diffusion model which contains two processes: diffusion process and reverse process. 
Instead of predicting the HR image directly, we apply residual prediction to predict the difference between the HR image $x_H$ and the upsampled LR image $up(x_L)$ and denote the difference as input residual image $x_0$. Diffusion process converts the $x_0$ into a latent $x_T$ in Gaussian distribution by gradually adding Gaussian noise $\epsilon$ as implied in Eq. \eqref{eq:x_arbitrary_t_on_x0}. 
According to Eq. \eqref{eq:reverse} and \eqref{eq:parameterization}, the reverse process is determined by $\epsilon_\theta$, which is a conditional noise predictor with an RRDB-based \cite{wang2018esrgan} low-resolution encoder (LR encoder for short) $\mathcal{D}$, as shown in Figure \ref{fig:conditional_noise_predictor}. 
The reverse process converts a latent variable $x_T$ to a residual image $x_r$ by iteratively denoising in finite step $T$ using the conditional noise predictor $\epsilon_\theta$, conditioned on the hidden states encoded from LR image by the LR encoder $\mathcal{D}$. The SR image is reconstructed by adding the generated residual image $x_r$ to the upsampled LR image $up(x_L)$. 
Therefore, the goal of $\epsilon_\theta$ is to predict the noise $\epsilon$ added at each diffusion timestep in the diffusion process\footnote{Instead of the original $L_2$ in Eq. \eqref{eq:train_obj}, we use $L_1$ for better training stability following Chen \etalcite{chen2020wavegrad}.}. 

In the following subsections, we will introduce architectures of the conditional noise predictor, LR encoder, training and inference.

\paragraph{Conditional Noise Predictor} 

The conditional noise predictor $\epsilon_\theta$ predicts noise added in each timestep of the diffusion process conditioned on the LR image information, according to Eq. \eqref{eq:train_obj} and \eqref{eq:parameterization}. As shown in Figure \ref{fig:conditional_noise_predictor}, we use U-Net as the main body, taking 3-channel $x_t$, the diffusion timestep $t\in\{1,2,...,T-1,T\}$ and the output of LR encoder as inputs. First, $x_t$ is transformed to hidden through a 2D-convolution block which consists of one 2D-convolutional layer and Mish activation \cite{misra2019mish}. Then the LR information is fused with the 2D-convolution block output hidden. Following Ho \etalcite{ho2020denoising}, we transform the timestep $t$ to the timestep embedding $t_e$ using the Transformer sinusoidal positional encoding \cite{vaswani2017attention}. Then the last output hidden and $t_e$ are fed into the contracting path, one middle step and the expansive path successively. The contracting path and expansive path both consist of four steps, each of which successively applies two residual blocks and one downsampling/upsampling layer. To reduce the model size, we only double the channel size in the second and the fourth contracting steps and halve the spatial sizes of the feature map in each contracting step. The downsampling layer in contracting path is a two-stride 2D convolution and the upsampling layer in expansive path is 2D transposed convolution. The middle step consists of two residual blocks, which is inserted between the contracting and expansive paths. Besides, the inputs of each expansive step concatenate the corresponding feature map from the contracting path. Finally, a 2D-convolution block is applied to generate $\hat{\epsilon}$ in timestep $t-1$ as the predicted noise, which is then used to recover $x_{t-1}$ according to Eq. \eqref{eq:reverse} and \eqref{eq:parameterization}. Our conditional noise predictor is easy and stable to train due to the multi-scale skip connection. Moreover, it combines local and global information through the contracting and expansive path.

\paragraph{LR Encoder} 
An LR encoder encodes the LR information $x_e$, which is added to each reverse step to guide the generation to the corresponding HR space. In this paper, we choose the RRDB architecture following SRFlow \cite{lugmayr2020srflow}, which employs the residual-in-residual structure and multiple dense skip connections without batch normalization layers. In particular, we abandon the last convolution layer of the RRDB architecture because we do not aim at the concrete SR results but the hidden LR image information. 

\paragraph{Training} 
In the training phase, as illustrated in Algorithm \ref{alg: training}, the input LR-HR image pairs in the training set are used to train SRDiff with the total diffusion step $T$ (Line 1). We randomly initialize the conditional noise predictor $\epsilon_\theta$ and the RRDB based LR encoder $\mathcal{D}$ is pretrained by $L_1$ loss function (Line 2). We then sample a mini-batch of LR-HR image pairs from the training set (Line 4) and compute the residual image $x_r$ (Line 5). The LR images are encoded by the LR encoder as $x_e$ (Line 6), which is fed into the noise predictor $\epsilon_\theta$ together with $t$ and $x_T$. Then we sample $\epsilon$ from the standard Gaussian distribution and $t$ from the integer set $\{1,\cdots,T\}$ (Line 7). We optimize the noise predictor by taking gradient step on Eq. (6) (Line 8).

\paragraph{Inference}
A T-step SRDiff inference takes an LR image $x_L$ as input (Line 1), as illustrated in Algorithm \ref{alg: sampling}. We sample a latent variable $x_T$ from the standard Gaussian distribution (Line 3) and upsample $x_L$ with bicubic kernel (Line 4). Different from the training procedure, we encode the LR image $x_L$ to $x_e$ by the LR encoder only once before the iteration begins (Line 5) and apply it in every iteration, which speeds up the inference. The iterations start from $t=T$ (Line 6), and each iteration outputs a residual image with a different noise level, which gradually declines as $t$ decreases. For $t>1$, we sample $z$ from standard Gaussian distribution (Line 7) and compute $x_{t-1}$ using the noise predictor $\epsilon_\theta$ with $x_t$, $x_e$ and $t$ as inputs (Line 8). Then for $t=1$, we set $z=0$ (Line 6) and $x_0$ is the final residual prediction (Line 10). An SR image is recovered by adding the residual image $x_0$ on the upsampled LR image $up(x_L)$.

\section{Experiments}
In this section, we first describe the experimental settings including datasets, model configurations and details in training and inference. Then we report experimental results and conduct some analyses.

\subsection{Experimental Settings}
\label{sec:exp_setting}
\paragraph{Datasets}
SRDiff is trained and evaluated on face SR ($8\times$) and general SR ($4\times$) tasks. For face SR, we use CelebFaces Attributes Dataset (CelebA)~\cite{liu2015CelebA}, which is a large-scale face attributes dataset with more than 200K celebrity images. The images in this dataset cover large pose variations and background clutter. In this paper, we use the whole training set which consists of 162,770 images for training and evaluate on 5000 images from the test split following SRFlow~\cite{lugmayr2020srflow}. We central-crop the aligned patches\footnote{\url{https://drive.google.com/drive/folders/0B7EVK8r0v71pWEZsZE9oNnFzTm8}} and resize them to $160\times160$ as HR ground truth using standard MATLAB bicubic kernel. To obtain the corresponding LR images, we downsample the HR images with bicubic kernel. For ProgFSR \cite{kim2019progressive}, we use progressive bilinear kernel introduced in its original paper for a fair comparison.

For General SR, we use the DIV2K~\cite{timofte2018ntire} and Flickr2K~\cite{timofte2018ntire}. These datasets consist of high-resolution RGB images with a large diversity of contents. In training, we use the whole training data (800 images) in DIV2K and whole images in Flickr2K (2650 images). Then we crop each image into patches with a size of $160\times160$ as HR ground truth following SRFlow. To obtain the corresponding LR images, we downsample the HR images with bicubic kernel. For evaluation, we use the whole validation data (100 images) in DIV2K. We downsample the HR images with bicubic kernel to obtain the LR images and directly apply SISR methods on the LR images to obtain the SR predictions without cropping.

\paragraph{Model Configuration}
Our SRDiff model consists of a 4-step conditional noise predictor and an LR encoder with multiple RRDB blocks. The number of channels $c$ in the first contracting step is set to 64. The numbers of RRDB blocks in the LR encoder are set to 8 and 15 for CelebA and DIV2K respectively and the channel size is set to 32. For diffusion process and reverse process, we set the diffusion step $T$ to $100$ and our noise schedule $\beta_1,...,\beta_T$ follows Nichol \etalcite{nichol2021improved}, which is proved to be beneficial for training. We also explore the model performance under different $T$ and $c$ in Section \ref{sec:ablation}.

\paragraph{Training and Evaluation}
Firstly, we pretrain the LR encoder $\mathcal{D}$ using an $L_1$ loss for 100k iterations for the sake of efficiency. The training of the conditional noise predictor uses Eq. \eqref{eq:train_obj} as loss term and Adam \cite{kingma2014adam} as optimizer, with batch size $16$ and learning rate $2 \times 10^{-4}$, which is halved every $100$k steps. The entire SRDiff takes about 34/45 hours (300k/400k steps) to train on 1 GeForce RTX 2080Ti with 11GB memory for CelebA/DIV2K respectively.

Beside the well-known evaluation metrics PSNR and SSIM \cite{wang2004image}, we also evaluate our SRDiff on LPIPS \cite{zhang2018unreasonable}, LR-PSNR \cite{lugmayr2020srflow} and the pixel standard deviation $\sigma$. LPIPS is recently introduced as a reference-based image quality evaluation metric, which computes the perceptual similarity between the ground truth and the SR images. LR-PSNR is computed as the PSNR between the downsampled SR image and the LR one indicating the consistency with the LR image. The pixel standard deviation $\sigma$ indicates diversity in the SR output.

\subsection{Performance}
In this subsection, we evaluate SRDiff by comparing with several state-of-the-art SR methods on face SR (8$\times$) and general SR ($4\times$) tasks. The detailed configurations of baseline models can be found in their original papers. %

\begin{table}[h]
\center
\small
\begin{tabular}{C{0.28cm} | C{1.3cm} | C{0.8cm} C{0.8cm} C{0.8cm} C{0.8cm} C{0.7cm}}
\toprule
& Methods & $\uparrow$PSNR & $\uparrow$SSIM & $\downarrow$LPIPS & $\uparrow$LR-PSNR & $\uparrow\sigma$ \\ 
\midrule
\multirow{5}{*}{\rotatebox[origin=c]{90}{\textit{Bicubic}}} 
& Bicubic  & 23.38 & 0.65  & 0.484 & 34.66 & 0.00 \\
& RRDB                          & 26.89 & 0.78  & 0.220 & 48.01 & 0.00 \\ %
& ESRGAN                        & 23.24 & 0.66  & 0.115 & 39.91 & 0.00\\ %
& SRFlow                        & 25.32 & 0.72  & 0.108 & 50.73 & 5.21 \\
& \textbf{SRDiff}               & 25.38 & 0.74  & 0.106 & 52.34 & 6.13 \\
\midrule
\multirow{3}{*}{\rotatebox[origin=c]{90}{\textit{Prog.}}} 
& ProgFSR         & 24.21 & 0.69  & 0.126 & 42.19 & 0.00 \\
& SRFlow          & 25.28 & 0.72  & 0.109 & 51.15 & 5.32 \\
& \textbf{SRDiff} & 25.32 & 0.73  & 0.106 & 51.41 & 6.19 \\
\bottomrule
\end{tabular}
\caption{Results for 8× SR of faces on CelebA. The first column indicates how LR images degenerate from HR ones and \textit{Prog.} means the progressive linear kernel from ProgFSR.}
\label{tab:celeba_results}
\end{table}

\paragraph{Face SR}
We compare SRDiff with RRDB \cite{wang2018esrgan}, ESRGAN \cite{wang2018esrgan}, ProgFSR \cite{kim2019progressive} and SRFlow ($\tau=0.8$) \cite{lugmayr2020srflow}\footnote{Due to inconsistent patch size, we retrain all these baseline models from scratch on our pre-processed CelebA dataset with released codes. SRFlow uses the same patch size as our model, but we cannot obtain the same patch with its released image example, and therefore, we also have to re-train it.}. RRDB is trained by only $L_1$ loss and can be regarded as a PSNR-oriented method. The evaluation results are shown in Table \ref{tab:celeba_results}, which reveals that SRDiff outperforms previous works in term of most of the evaluation metrics, and can generate high-quality and diverse SR images with strong LR-consistency. Specifically, 1) as shown in Figure \ref{fig:faces8x}, compared with PSNR-oriented methods, SRDiff reconstructs clearer textures, and compared with GAN-driven methods, SRDiff avoids artifacts and the results look more natural; and 2) as shown in Figure \ref{fig:faces_diverse}, SRDiff provides diverse and realistic SR predictions given only one LR input. Every SR prediction is a complete portrait of a human face with rich details and maintains consistency with the input LR image. Moreover, SRDiff uses fewer model parameters (12M) than SRFlow (40M) and only takes about 30 hours until converge as described in Section \ref{sec:exp_setting}, while SRFlow needs 5 days, which demonstrates that SRDiff is training-efficient and can achieve comparable performance with a small model footprint since SRDiff does not impose any architectural constraints to guarantee bijection. Compared with GAN-driven methods, SRDiff does not need any extra module (e.g., discriminator) in training. 

\begin{table}[h]
\center
\small
\begin{tabular}{C{1.8cm} | C{0.8cm} C{0.8cm} C{0.8cm} C{0.8cm} C{0.7cm}}
\toprule
Methods & $\uparrow$PSNR & $\uparrow$SSIM & $\downarrow$LPIPS & $\uparrow$LR-PSNR & $\uparrow\sigma$ \\ 
\midrule
Bicubic & 26.70 & 0.77 & 0.409 & 38.70 & 0.00  \\
EDSR & 28.98 & 0.83 & 0.270 & 54.89 & 0.00   \\
RRDB & 29.44 & 0.84 & 0.253 & 49.20 & 0.00  \\ 
RankSRGAN & 26.55 & 0.75 & 0.128 & 42.33 & 0.00  \\ 
ESRGAN & 26.22 & 0.75 & 0.124 & 39.03 & 0.00  \\
SRFlow & 27.09 & 0.76 & 0.120 & 49.96 & 5.14  \\
\textbf{SRDiff} & 27.41 & 0.79 & 0.136 & 55.21 & 6.09  \\
\bottomrule
\end{tabular}
\caption{Results for 4× SR of general images on DIV2K.}
\label{tab:div2k_results}
\end{table}

\paragraph{General SR}
We also evaluate SRDiff on general SR ($4\times$) compared with EDSR \cite{lim2017enhanced}, RRDB, ESRGAN, RankSRGAN \cite{zhang2019ranksrgan} and SRFlow ($\tau=0.9$) with their official released pretrained models\footnote{Except RRDB which is trained from scratch with $L_1$ loss as that in Face SR.}. As shown in Table \ref{tab:div2k_results}, SRDiff achieves better quantitative results than the previous methods for most evaluation metrics (PSNR, SSIM and LR-PSNR) and comparable LPIPS, which reveals the effectiveness and great potential of our method. Figure \ref{fig:general4x} shows that SRDiff balances sharpness and naturalness well and produces strong consistency with the LR image. In contrast, PSNR-oriented methods (EDSR and RRDB) and SRFlow, smear the edges of the objects, and GAN-driven methods (ESRGAN and RankSRGAN) introduce more artifacts.

\begin{figure*}[!t]
	\centering
	\begin{subfigure}[h]{0.40\textwidth}
		\centering
			\includegraphics[width=\textwidth,trim={0cm 0cm 0.45cm 0cm}, clip=true]%
		{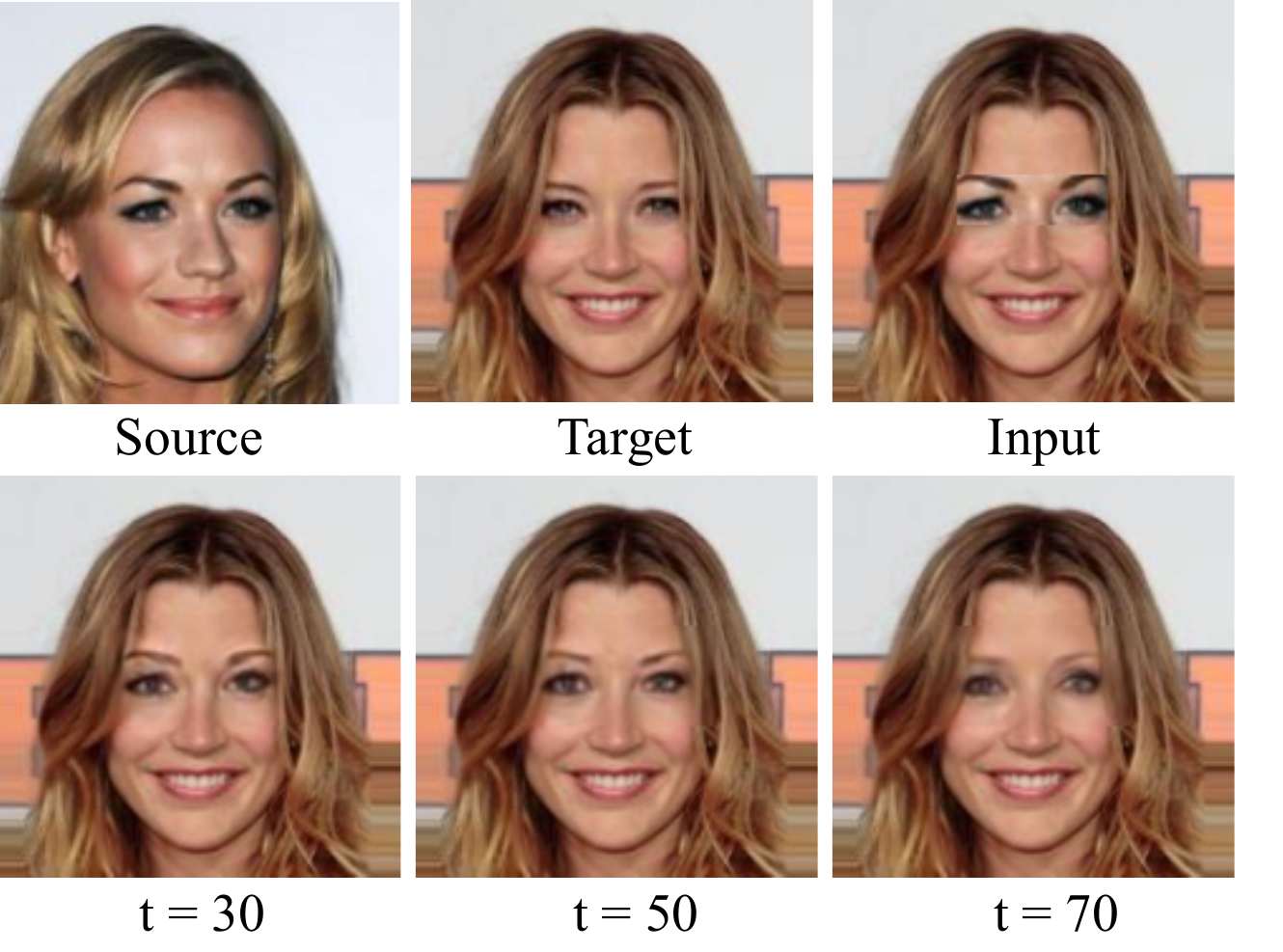}
    	\vspace{-0.4cm}
    	\caption{SRDiff applying on content fusion task.}
    	\label{fig:content_transfer}
	\end{subfigure}
	\hfill
	\begin{subfigure}[h]{0.55\textwidth}
	    \vspace{0.2cm}
	    \includegraphics[width=\textwidth,trim={0cm 0.5cm 0.4cm 0cm}, clip=true] %
	{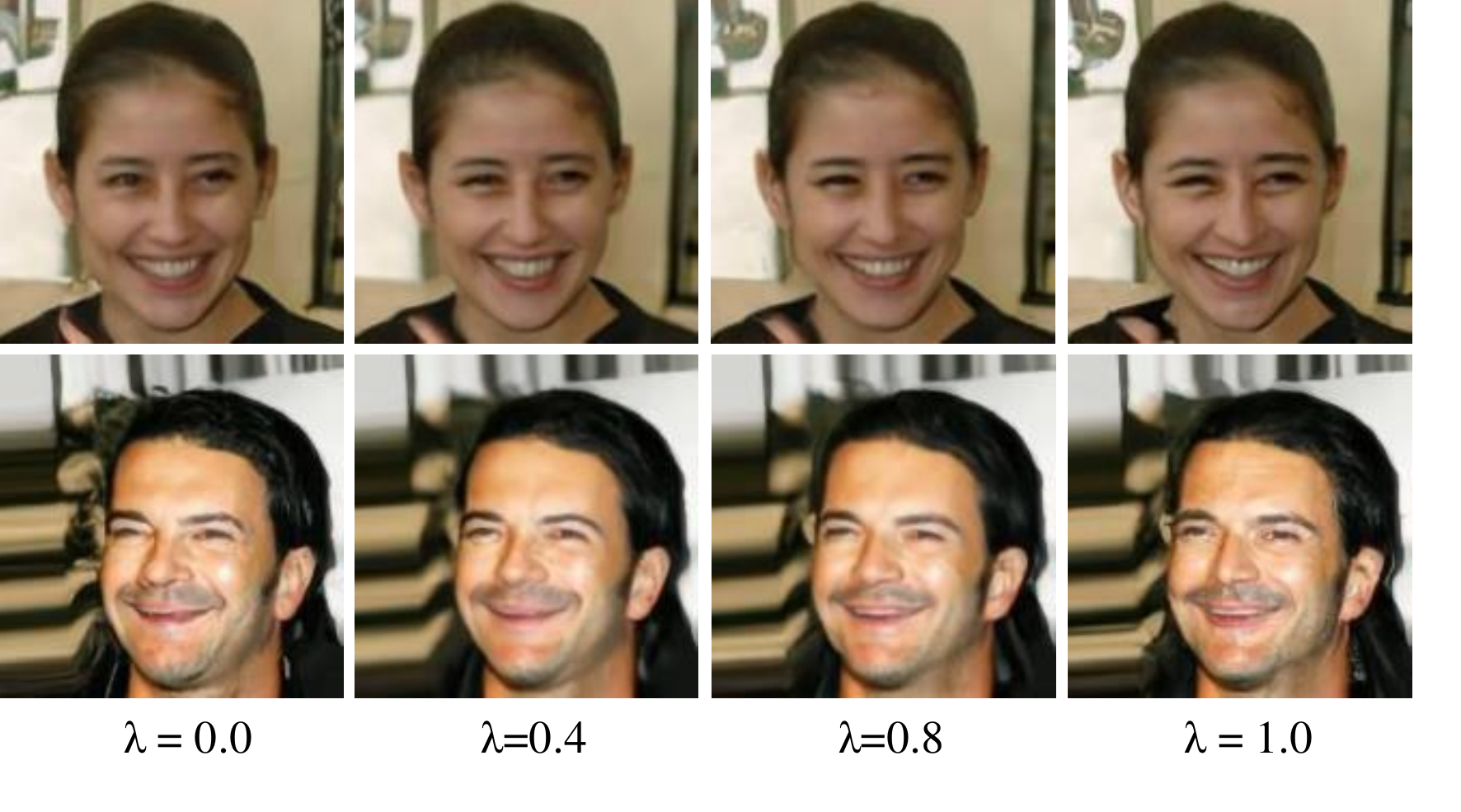}
	    \vspace{-0.45cm}
	    \caption{SRDiff applying on latent interpolation task.}
	    \label{fig:latent_interpolation}
	\end{subfigure}
	\caption{Extended applications of SRDiff.}
	\label{fig:ext_app}
\end{figure*}

\begin{table}[h]%
\small
\center
\begin{tabular}{C{0.7cm} C{0.5cm} C{0.6cm} | C{0.73cm} C{0.73cm} C{0.73cm} C{0.73cm} C{0.7cm}}
\toprule
$T$ & $c$ & \textit{Res.} & $\uparrow$PSNR & $\uparrow$SSIM & $\downarrow$LPIPS & $\uparrow$LR-PSNR & $\downarrow$\textit{Steps} \\ 
\midrule
100 & 64 & $\surd$ & 25.38 & 0.74  & 0.106 & 52.34 & 300k \\
\midrule
25 & 64 & $\surd$ & 25.12 & 0.71  & 0.109 & 52.17 & 300k \\
200 & 64 &$\surd$  & 25.41 & 0.74 & 0.106 & 52.31 & 300k \\
1000 & 64 &$\surd$  & 25.43 & 0.75 & 0.105 & 52.35 & 300k \\
100 & 32 &$\surd$  & 25.15 & 0.72 & 0.108 & 52.20 & 300k \\
100 & 128 &$\surd$  & 25.40 & 0.74 & 0.106 & 52.37 & 300k \\
100 & 64 & $\times$ & 24.88 & 0.70 & 0.111 & 51.90 & 600k \\
\bottomrule
\end{tabular}
\caption{Ablations of SRDiff for faces SR ($8\times$) on CelebA. $T$, $c$ and $\textit{Res.}$ denote the total diffusion step, channel size of the noise predictor, and the residual prediction respectively.}
\label{tab:abalation_results}
\end{table}

\subsection{Ablation Study}
\label{sec:ablation}
To probe the influences of the total diffusion step $T$, noise predictor channel size $c$ and the effectiveness of residual prediction, we conduct some ablation studies as illustrated in Table \ref{tab:abalation_results}. From row 1, 2, 3 and 4, we can see that the image quality is enhanced as total diffusion steps increases. From row 1, 5 and 6, we can see that larger model width results in better performance. However, larger total diffusion steps and model width both lead to slower inference, and therefore, we choose $T=100$ and $c=64$ as the default setting after trading off. Row 1 and 7 indicate that the residual prediction not only greatly enhances the image quality but also speeds up the training, which demonstrates the effectiveness of residual prediction.

\subsection{Extensions}
In this subsection, we explore some extended applications including content fusion and latent space interpolation. %
\paragraph{Content Fusion}
SRDiff is applicable in content fusion tasks, which aim to generate an image by fusing contents from two source images, e.g., an eye-source image and a face-source image providing the eye and face contents respectively. In this paragraph, we use SRDiff to conduct face content fusion by a demonstration of fusing one's eyes with another one's face. The procedure of content fusion is shown in Algorithm \ref{alg:content_fusion}. First, we directly fuse a face image $x_{f}$ by replacing the eye region of the face-source image $x_{face}$ with that of the eye-source image $x_{eye}$ and compute the differences between $x_{f}$ and the upsampled LR face-source image $up(x_{L})$ to get the residual $x_r$. Second, $x_r$ goes through a $\bar{t}$-step diffusion process, which outputs the $x_{\bar{t}}$ in latent space. Then $x_{\bar{t}}$ is denoised to an HR residual using the conditional noise predictor iterated from $\bar{t}$ to $0$ with the LR face-source information encoded by the LR encoder, which ensures the compatibility of the two contents. Then we get the fused SR image by adding the SR residual to $up(x_L)$. Finally, we replace the eye region of the face-source image with that of the SR face image and preserve the non-manipulated face. As shown in Figure \ref{fig:content_transfer}, we set different timesteps $t \in \{30,50,70\}$ and find that the eye region of the fusion result is more similar to the eye-source image when $t$ is small, and is closer to the face-source image as $t$ becomes larger.

\setcounter{algorithm}{2}
\begin{algorithm}[H] 
    \centering 
    \small
    \caption{Content Fusion}\label{alg:content_fusion} 
    \begin{algorithmic}[1] 
        \STATE\textbf{Input}: eye-source image $x_{eye}$, face-source image $x_{face}$, diffusion step $\bar{t}$.
        \STATE\textbf{Load}: conditional noise predictor $\epsilon_\theta$ and LR encoder $\mathcal{D}$.
        \STATE Replace the eye region of $x_{face}$ by the that of $x_{eye}$ to form the initial fused image $x_f$
        \STATE Upsample the LR face-source image $x_L$ as $up(x_L)$
        \STATE Compute $x_r=x_f-up(x_L)$ 
        \STATE Put $x_r$ into the diffusion process and compute $x_{\bar{t}}(x_r, \epsilon) = \sqrt{\bar\alpha_{\bar{t}}}x_r + \sqrt{1-\bar\alpha_{\bar{t}}}\epsilon,~ \epsilon \sim \mathcal{N}(\textbf{0}, \textbf{I})$
        \STATE Encode $x_L$ as $x_e = \mathcal{D}(x_L)$ 
        \FOR{$t=\bar{t},\cdots,1$} 
            \STATE Sample $z \sim \mathcal{N}(\textbf{0}, \textbf{I})$ if $t > 1$, else $z = 0$ 
            \STATE Compute $x_{t-1} = \frac{1}{\sqrt{\alpha_t}}\left(x_t-\frac{1-\alpha_t}{\sqrt{1-\bar\alpha_t}} \epsilon_\theta(x_t,x_e,t)\right) + \sigma_\theta(x_t, t) z$ 
        \ENDFOR 
        \STATE Compute the SR face prediction $x_H=x_0+up(x_L)$
        \STATE Crop the eye region of $x_H$ and insert it to the corresponding eye region of $x_{face}$ to generate $x_{fused}$
        \RETURN{$x_{fused}$} as the content fusion result
    \end{algorithmic} 
\end{algorithm} 

\paragraph{Latent Space Interpolation}
Given an LR image, SRDiff can manipulate its prediction by latent space interpolation, which linearly interpolates the latents of two SR predictions and generate a new one. Let $x_{\bar{t}},x'_{\bar{t}}\sim q(x_{\bar{t}}|x_0)$ and we decode the latent $\bar{x}_{\bar{t}} = \lambda x_{\bar{t}} + (1-\lambda)x'_{\bar{t}}$ by the reverse process, which feeds $\bar{x}_{\bar{t}}$ into the noise predictor with the LR information encoded by LR encoder iteratively. Then, we add the output residual result to the $up(x_{L})$ to obtain the interpolated SR prediction. We set $\bar{t}=50$ and $\lambda \in \{0.0,0.4,0.8,1.0\}$. It could be observed in Figure \ref{fig:latent_interpolation} that with $\lambda$ approaching to 1.0, the woman's expression becomes closer to $x_t$, which is the top right image holding a big laugh. In the same way, the man's mouth turns wider and bigger from $\lambda=0.0$ to $\lambda=1.0$. The trend of the interpolated images shows the effectiveness of SRDiff in latent space interpolation. The detailed algorithm of latent space interpolation is shown in Algorithm \ref{alg:latent_space_interpolation}.

\setcounter{algorithm}{3}
\begin{algorithm}[H] 
    \centering 
    \small
    \caption{Latent Space Interpolation}\label{alg:latent_space_interpolation} 
    \begin{algorithmic}[1] 
        \STATE\textbf{Input}: LR image $x_L$, diffusion step $\bar{t}$, $\lambda \in[0,1]$
        \STATE\textbf{Load}: conditional noise predictor $\epsilon_\theta$ and LR encoder $\mathcal{D}$
        \STATE Sample $x_{\bar{t}},x_{\bar{t}}'~\sim \mathcal{N}(\textbf{0}, \textbf{I})$
        \STATE Compute $\bar{x_t} = \lambda x_{\bar{t}} + (1-\lambda)x_{\bar{t}}'$
        \STATE Upsample $x_L$ as $up(x_L)$
        \STATE Encode $x_L$ as $x_e = \mathcal{D}(x_L)$ 
        \FOR{$t=\bar{t},\cdots,1$} 
            \STATE Sample $z \sim \mathcal{N}(\textbf{0}, \textbf{I})$ if $t > 1$, else $z = 0$ 
            \STATE Compute $x_{t-1} = \frac{1}{\sqrt{\alpha_t}}\left(x_t-\frac{1-\alpha_t}{\sqrt{1-\bar\alpha_t}} \epsilon_\theta(\bar{x_t},x_e,t)\right) + \sigma_\theta(x_t, t) z$ 
        \ENDFOR 
        \RETURN the interpolated SR face prediction $x_H=x_0+up(x_L)$ as the latent interpolation results
    \end{algorithmic} 
\end{algorithm} 

\section{Conclusion}
In this paper, we proposed SRDiff, which is the first diffusion-based model for single image super-resolution to the best of our knowledge. Our work exploits a Markov chain to convert HR images to latents in simple distribution and then generate SR predictions in the reverse process which iteratively denoises the latents using a noise predictor conditioned on LR information encoded by the LR encoder. To speed up convergence and stabilize training, SRDiff introduces residual prediction. Our extensive experiments on both face and general datasets demonstrate that SRDiff can generate diverse and realistic SR images and avoids over-smoothing and mode collapse issues that occurred in PSNR-oriented methods and GAN-driven methods respectively. Moreover, SRDiff is stable to train with small footprint and without an extra discriminator. Besides, SRDiff allows for flexible image manipulation including latent space interpolation and content fusion. 

In the future, we will further improve the performance of the diffusion-based SISR model and speed up the inference. We will also extend our work to more image restoration tasks (e.g., image denoising, deblurring and dehazing) to verify the potential of diffusion models in the image restoration domain. 

\bibliographystyle{named}
\bibliography{ijcai21}

\end{document}